\documentclass[12pt,compsoc]{IEEEtran}
\usepackage{lipsum}

\usepackage{authblk}
\usepackage{amssymb}
\usepackage{amsmath}
\usepackage{mathrsfs}
\usepackage{amsfonts}
\usepackage{braket}
\usepackage{color}
\usepackage{graphicx}
\usepackage{tikz}
\usepackage{pgf}
\usetikzlibrary{arrows,automata,positioning,decorations.pathreplacing,shapes,matrix}
\usepackage[english]{babel}
\newcommand{\abs}[1]{\left\vert#1\right\vert}

\def\XXint#1#2#3{{\setbox0=\hbox{$#1{#2#3}{\int}$ }
\vcenter{\hbox{$#2#3$ }}\kern-.6125\wd0}}

\newcounter{lastnote}

\begin{document}
\title{Bayesian Networks based Hybrid Quantum-Classical Machine Learning Approach to Elucidate Gene Regulatory Pathways}
\author{$^\dag$Radhakrishnan Balu and $^\ddag$Ajinkya Borle}
\affil{$^\dag$Army Research Laboratory Adelphi, MD, 21005-5069, USA \\
       radhakrishnan.balu.civ@mail.mil } 
\affil{$^\ddag$Computer Science and Electrical Engineering, \\
       University of Maryland Baltimore County, \\
      1000 Hilltop Circle, Baltimore, MD 21250}          
\date{Received: date / Accepted: \today}
\IEEEtitleabstractindextext{%
    \begin{abstract}
We report a scalable hybrid quantum-classical machine learning framework to build Bayesian networks (BN) that captures the conditional dependence and causal relationships of random variables. The generation of a BN consists of finding a directed acyclic graph (DAG) and the associated joint probability distribution of the nodes consistent with a given dataset. This is a combinatorial problem of structural learning of the underlying graph, starting from a single node and building one arc at a time, that fits a given ensemble using maximum likelihood estimators (MLE). It is cast as an optimization problem that consists of a scoring step performed on a classical computer, penalties for acyclicity and number of parents allowed constraints,  and a search step implemented using a quantum annealer. We have assumed uniform priors in deriving the Bayesian network that can be relaxed by formulating the problem as an estimation Dirichlet parameters. We demonstrate the utility of the framework by applying to the problem of elucidating the gene regulatory network for the MAPK/Raf pathway in human T-cells using proteomics data where the concentration of proteins, nodes of the BN,  are interpreted as probabilities. 
\end{abstract}
}
\maketitle
\IEEEdisplaynontitleabstractindextext
\section{Introduction}
\label{intro}
Determining the causal relationship of a biochemical pathway is crucial to develop strategies for therapeutic interventions when the pathways involved are the basis for molecular physiologies. Figure 1 depicts the set of proteins that are activated as a result of extracellular signals in MAPK/Ras pathway similar to the one considered in this study. The causations are distinguished from associations of involved proteins through randomized experiments and this problem can be cast as a Bayesian network determination that closely models the statistics of the observed protein concentrations interpreted as probabilities. The proteins are the nodes of the BN and there is an arc in the directed acyclic graph when there is a conditional dependence between a node and its parents. 
\begin{figure*}[!b]
 \begin {center}
\includegraphics[width=1\columnwidth]{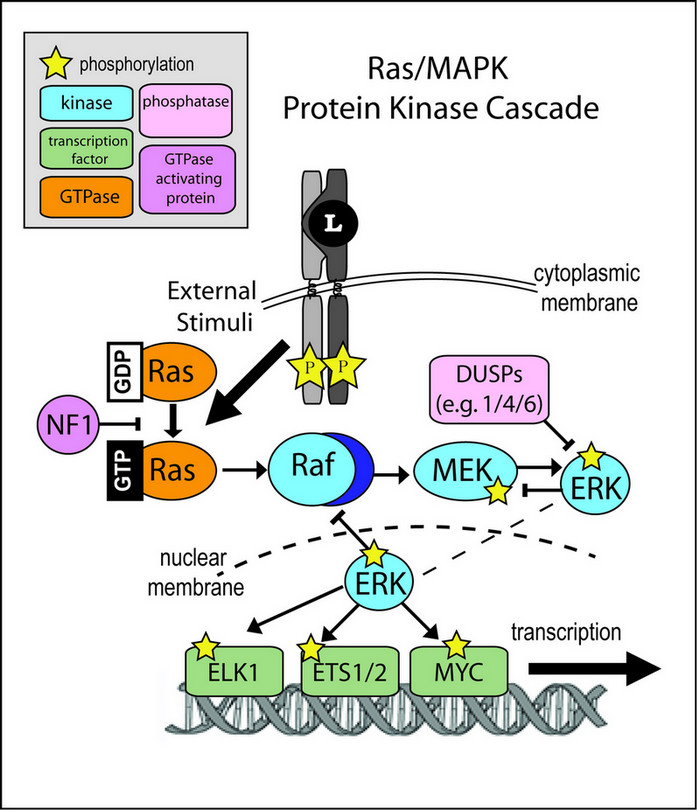}
\end{center}
\caption {MAPK/Ras pathway. Source: DiscoveryMedicine}
\end {figure*}[!b]

Starting from the raw data of protein concentrations the probabilities of the nodes are assigned by converting them in to normalized probabilities. Then, deriving the most likely underlying BN is a combinatorial problem exploring the space of all possible arcs between the nodes satisfying constraints such as acyclicity, maximum number of parents that could be allowed for a node, and directionality of the dependencies consistent with the probabilities. The construction of the underlying graph one arc at a time can be formulated as an energy minimization problem by choosing an appropriate score, that involve Dirichlet priors, for different configurations. Further, the priors themselves can be estimated using a machine learning paradigm with a training set. 

In this work we developed an hybrid quantum-classical solution to the derivation of Bayesian networks specialized for modeling biochemical pathways with the encoding of Dirichlet priors estimation and domain specific knowledge using a probabilistic logic programming language and the construction of BN graph as a quantum annealing process  \cite{Kadowaki1998} using the D-Wave II architecture. In this work we present the results of applying the framework to the MAPK/Raf signal transduction pathway \cite {Sachs2005}. We have assumed an uniform distribution for the Dirichlet priors and in the future will implement the learning associated with the parameter estimation.

Bayesian networks are formally defined as a pair $(B_S, B_P)$ with the first component describing a directed acyclic graph (DAG) whose nodes are random variables, $\{X_i, 1\leq{i}\leq{n}\}$, and the second part is a conditional probability of each variable given that of its parents, $\{p(X_{i}|\Pi(X_i))\}$. Given a set of data D of N instances of the state $\{X_i, 1\leq{i}\leq{n}\}$ (ensemble) the goal is to construct the most probable structure, that is the $B_S$ that maximizes the posterior probability $p(B_s|D)$. From Bayes theorem we have this probability proportional to $p(D|B_S)$ as
\begin {equation}
p(B_S|D) = \frac{p(D|B_S)p(B_S)}{p(D)}.
\end {equation}
The above conditional probability is expressed in terms of the hyper parameters $\alpha_{ijk}$ of the Dirichlet priors, typically assumed to be uniformly distributed, and pseudocounts $N_{ijk}$ as
\begin {equation} \label{eq:likelihood}
p(D|B_S) = \Pi_{i=1}^n\Pi_{j=1}^{q_i}\frac{\Gamma(\alpha_{ij})}{\Gamma(N_{ij} + \alpha_{ij})}\Pi_{k=1}^{r_i}\frac{\Gamma(N_{ijk} + \alpha_{ijk})}{\Gamma(\alpha_{ijk})}.
\end {equation}
The discrete probability distribution of r.v. $X_i$ is denoted by $\theta_{ij},1\leq{j}\leq{r_i}$ and $\pi_{ij}$ refers to j-th joint state of the parent set $\Pi_i$. The conditional probability $p(x_{ik}|\pi_{ij})$ of the r.v. $X_i$ in the state k given the j-th joint  distribution of its parents is denoted by $\theta_{ijk}$.

Quantum annealing (QA) is an optimization process where the objective function defines a penalty function for each unsatisfied clause, with the ultimate goal of finding the truth assignment that minimizes the objective function. This process starts with an easy to prepare quantum ground state, and the system is evolved adiabatically: the system remains in its ground state at all times, to its energy minimum state which is interpreted as the solution to a given problem \cite{FA}. The adiabatic evolution is described by a Hamiltonian with two parts, the driver referred as $\text{H}_{B}$ and the second part denoted as  $\text{H}_{P}$ to designate the problem Hamiltonian. During the evolution, the contribution from $\text{H}_{B}$ is slowly decreased from 100\% while that of $\text{H}_{P}$ is increased from zero to 100\%.  One of the key issues in designing quantum adiabatic algorithms is to ensure a sufficient gap exists between the ground and first excited states of the Hamiltonian to prevent a phase transition \cite{SAT2, SAT3}. The adiabatic evolution time has to be sufficiently long so a fixed point or an approximate equilibrium point is reached. In several situations an approximate equilibrium point would be adequate. During the evolution it is important to prevent the system from reaching the excited state and thus the gap matters in achieving this goal. At the same time, it is desirable to have a bound on the gap so that the equilibrium state is reached in a reasonable time. Estimating the energy gap of an arbitrary Hamiltonian is an open problem that will not be addressed here and tools can be designed to calculate the gap for specific situations. The quantum annealer supported by the D-Wave II architecture enables development of solutions to wide range of problems in computer science \cite {Gaitan2014} and physics \cite {Guzik2008} that can be cast as quadratic unconstrained binary optimization (QUBO).   

\section {Results and Discussions}
We have set up our QUBO after the prescription of \cite {Alan2015} Aspuru-Guzik et. al work and extended it to treat BN with larger number of nodes and more relaxed conditions such as each node can at most three parents (the referenced work assumed at most two parents per node). There are n(n - 1) possible arcs in a graph with n nodes and we need so many bits to store the information.
We use the notation d for the set of all arcs of the graph where $d_{ij}, 1\leq{i,j}\leq{n},i\neq{j}$ represents the arc from node i to j that is a binary variable. 

It is convenient to work with the logarithmic form of the equation \eqref{eq:likelihood} which is in a separable form of the nodes.
\begin {multline}
s_i(\Pi_i(B_S)) = -log\left(\Pi_{j=1}^{q_i}\frac{\Gamma(\alpha_{ij})}{\Gamma(N_{ij} + \alpha_{ij})}\right)  \\
\left( \Pi_{k=1}^{r_i}\frac{\Gamma(N_{ijk} + \alpha_{ijk})}{\Gamma(\alpha_{ijk})}\right).
\end {multline}
The negative sign is to convert the problem in to a minimization as
\begin {equation}
log(p(D|B_S) = -s(B_s) = \sum_{i = 1}^n s_i(\Pi_i(B_S)).
\end {equation}
The Hamiltonian \cite {Alan2015} whose ground state encodes the solution to the BN that we want to construct for the given data is set to the log likelihood function. It is in a separable form of arcs originating from node i.
\begin {equation}
H_{score}(d) = \sum_{i = 1}^n H_{score}(d_i).
\end {equation}
The final scoring Hamiltonian with the assumption of at most three parents per node $\abs{J} \leq {3}$ as a weighted states on arcs is 
\begin {equation}
H_{score}^{(i)}(d_i) = \Sigma_{J\subset\{1,...,n\}\backslash{\{i\}}} \left( w_i(J) \Pi_{j\in{J}} d_{ji} \right).
\end {equation}
To enforce the maximum parents allowed per node constraint we need the following Hamiltonian:
\begin {align}
d_i &= \sum_{1\leq{j} \leq {n}} d_{ji}. \\
\mu &= \lceil log_2(m + 1) \rceil.\\
y_i &= \sum_{l = 1}^\mu 2^{l - 1} y_{il}. \\
H_{max}^{(i)}(d_i, y_i) &= \delta^{(i)}_{max} (m - d_i - y_i)^2.
\end {align}
In the above $\delta$ is the penalty to discourage solutions exceeding the limit, $d_i$ is the incoming degree of node i and $y_i \in \mathbb{Z}$ is an auxiliary variable for each node represented in binary form. In our case of maximal three parents DAGs we need two qubits per node. We used ancilla qubits to convert the three-body terms to two-body terms suitable for D-Wave supported QUBOs.
Finally, we need constraints for acyclicity that has two parts, a transitive terms that is zero when the relation encoded in $\{r_{ij}\}$ is zero only when it is indeed transitive and positive otherwise. The second part enforces consistency by the term evaluating to zero if the order encoded in $\{r_{ij}\}$ is in accordance with the directed edge structure $\{d_{ij}\}$ and positive otherwise. The corresponding Hamiltonians are:
\begin {multline}
H_{trans} = \sum_{1\leq{j}\leq{n}} H_{trans}^{ijk} (r_{ij}, r_{ik}, r_{jk}). \\
H_{trans}^{ijk} (r_{ij}, r_{ik}, r_{jk}) = \delta_{trans}^{(ijk)} \text{  if } \\
(x_i\leq{x_j}\leq{x_k}) \vee (x_i\geq{x_j}\geq{x_k}). \\
= \text{  0  o.t.  }\\
H_{consist} (d, r) = \sum_{1\leq{j}\leq{n}} H_{consist}^{ij} (d_{ij}, d_{ik}, d_{jk}). \\
H_{consist}^{ij} (d_{ij}, d_{ik}, d_{jk}) = \delta_{consist}^{(ij)} \\
\text{  if  } (d_{ji} = r_{ij} =1) \vee (d_{ij} = 1 \wedge r_{ij} = 0). \\
= \text{  0  o.t.  }\\
H_{cycle} = H_{trans} + H_{consist}.
\end {multline} 
\begin {multline}
H_{total} (d, y, r)= H_{score} (d) + H_{max} (d, y) + \\
H_{cycle} (d, r).
\end {multline}
The penalties used in different parts of the Hamiltonian need to be adjusted to steer the solution towards the low energy configuration which at this stage is still an art.
The original biological network has eleven proteins and lipids participating in the signal transduction and out of which three of them (PLCG, PIP3, and PIP2) are independent forming a clique in the BN graph. As a result we considered only eight of the proteins depicted in Figure 2. We used the data discretized \cite {Scutari2015} using a prescription that jointly discretize all the variables \cite {Hartemink2001} into a small number of intervals by iteratively collapsing the intervals defined by their quantiles.

Our goal in this study is to demonstrate the viability quantum annealing techniques supported by D-Wave II architecture for real world problems such as the signal transductions that are of paramount importance in pharmaceutical sciences. Our encoding of the problem involved  205 logical qubits that we managed to embed on the 2000 qubits processor utilizing all most all of the qubits. Mapping the logical qubits of the Hamiltonian onto the physical hardware a bipartite graph is one of the technical challenges part of the study as it is NP-hard forcing an heuristic solution. Our embedding could only accommodate three-parent constraint applied to seven of the nodes and only MEK protein has this additional requirement and so we we were able to construct the pathway. We used the R statistics package \cite {R2013} to calculate the scores from the digitized protein concentration in order to set up the Hamiltonian for the annealer. Our framework reproduces the original biological pathway qualitatively as in our constructions we mostly avoided cycles formations and realized three parents constraint as the MEK protein is activated by three different pathways. We have the results of the annealings of 30 schedules, 9000 runs per schedule, in Figure 3 for the MAPK/Raf pathway for which the causal relationship is captured by the Bayesian network in Figure 2. The number of false positives generated is also quite high due to the noisy nature of the qubits. We applied simple error correction techniques such as the gauge transformations that involve flipping the spins that didn't significantly improve the results. In the classical context the elucidation missed few arcs due to the acyclicity constraint \cite {Sachs2005} and our framework constructed most of the crucial arcs, taking into account the individual genetic variations in a population this is critical instead of getting all the arcs correct, without assuming any prior connectivity. In addition, the arcs were constructed with a sample of limited dataset and averaged over the population that masked some of the signals as in the classical computing approach. 

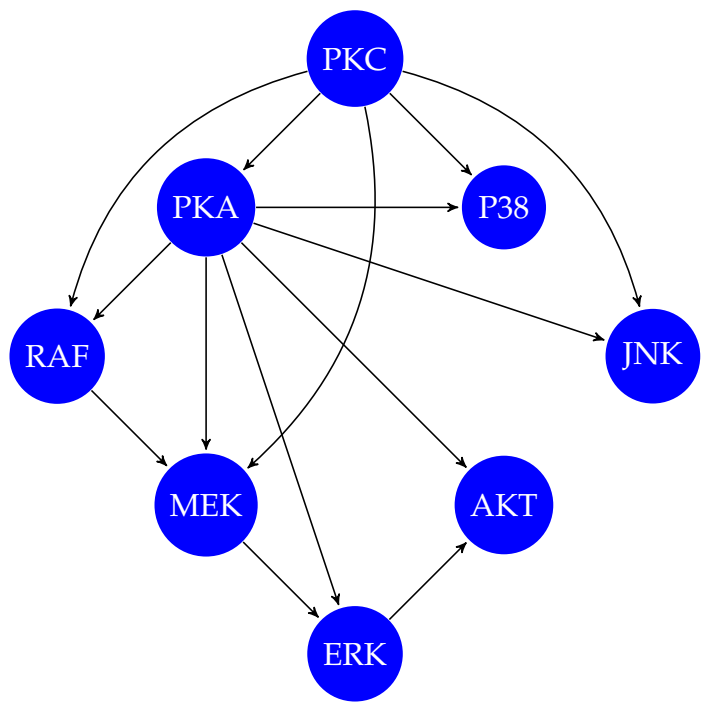
\begin{figure*}[!b]
\centering
\begin{tikzpicture}[->,>=stealth',shorten >=1pt,auto,node distance=2.8cm,
                    semithick]                   
  \tikzstyle{every state}=[fill=blue,draw=none,text=white]

  \node[state]         (A)                    {PKC};
  \node[state]         (B) [below left of=A] {PKA};
  \node[state]         (D) [below right of=A] {P38};
  \node[state]         (C) [below right of=D] {JNK};
  \node[state]         (E) [below left of = B]  {RAF};
   \node[state]        (F) [below right of=E] {MEK};
  \node[state]         (G) [below right of=F] {ERK};
  \node[state]         (H) [below left of=C]       {AKT};

  \path (A) edge              node  {} (B)
            edge    [bend left, right]                node  {} (C)
            edge                   node  {} (D)
            edge  [bend right, left]                 node  {} (E)
            edge  [bend left, right]                  node  {} (F)
        (B) edge                 node  {} (D)
              edge                 node  {} (C)
              edge                 node  {} (E)
              edge                 node  {} (F)
              edge                 node  {} (G)
              edge                 node  {} (H)
        (E) edge                 node {} (F)
        (F) edge                 node {} (G)
        (G) edge                 node {} (H);

\end{tikzpicture}
\caption {Bayesian Network encoding the causal relationship in MAPK/Raf signal transduction pathway in human T-cells. Only eight of the units are considered here.}
\centering
 \end {figure*}
 
 \newpage
 \tikzset{ 
    table/.style={
        matrix of nodes,
        row sep=-\pgflinewidth,
        column sep=-\pgflinewidth,
        nodes={
            rectangle,
            draw=black,
            align=center
        },
        minimum height=6.0em,
        text depth=0.5ex,
        text height=2ex,
        nodes in empty cells,
        every even row/.style={
            nodes={fill=gray!20}
        },
        column 1/.style={
            nodes={text width=6em,font=\bfseries}
        },
        row 1/.style={
            nodes={
                fill=black,
                text=white,
                font=\bfseries
            }
        }
    }
}

\begin{figure*}
\centering
\begin{tikzpicture}

\matrix (first) [table,text width=6em]
{
& Raf BN   & Raf BN with Error Correction\\
No of instances with cycles   & 0 & 1  \\
No of false positives    & 6-13 & 6-14  \\
Average true positives   & 8.3 & 8.6 \\
Median of true positives  & 9 & 9 \\
};
\end{tikzpicture}
 \caption {Results from 30 annealing schedules for the MAPK/Raf BN with 14 arcs.}
 \centering
 \end {figure*}

\section {Summary and conclusions}
We have developed a hybrid quantum-classical machine learning framework to derive a BN that model causal relationships between protein activation in a biological pathway. We demonstrated the viability of using the D-Wave annealer for real world applications by elucidating crucial features of the signal transduction and the framework is flexible enough to treat wide range of situations. We have assumed an uniform priors in constructing the BN and this can be relaxed by estimating the Dirichlet parameters of the distribution either using classical computation or by annealing approaches. Other extension we have planned include dynamical Bayesian Networks and Bayesian Neural Networks. With the advances in greater connectivity, increased scaling of qubits, and improved error correction planned in D-Wave systems larger pathways can be studied using this hybrid approach.
\section* {Acknowledgements}
The authors would like to thank the staff at D-Wave systems, Rene Copeland and Denny Dahl for providing access to the hardware and the technical support in using the system.

\newpage

\end{document}